\documentclass[sigconf, nonacm]{acmart}
\AtBeginDocument{%
  }

\setcopyright{acmlicensed}
\copyrightyear{2026}
\acmYear{2026}
\acmDOI{XXXXXXX.XXXXXXX}
\acmConference[KDD '26]{Proceedings of the 32nd ACM
SIGKDD Conference on Knowledge Discovery and Data Mining}{August 9--13,
  2026}{Jeju, Korea}

\usepackage{style}
\usepackage{amsmath}

\usepackage{amssymb}
\usepackage{mathtools}
\DeclareMathOperator*{\argmax}{arg\,max}
\usepackage{amsthm}
\usepackage{xcolor}
\usepackage{subcaption}
\usepackage{wrapfig}

\newcommand{\nbo}[1]{{\sf\color{orange}[#1]}}
\newcommand{\ls}[1]{\nbo{LS: #1}} 

\newcommand{\nbb}[1]{{\sf\color{blue}[#1]}}
\newcommand{\cm}[1]{\nbb{CM: #1}} 

\usepackage[capitalize,noabbrev]{cleveref}

\theoremstyle{plain}
\theoremstyle{definition}
\theoremstyle{remark}

\begin{document}

\title{Similarity-Aware Mixture-of-Experts for Data-Efficient Continual Learning}

\author{Connor Mclaughlin}
\affiliation{%
  \institution{Northeastern University}
  \city{Boston}
  \state{Massachusetts}
  \country{USA}
}
\author{Nigel Lee}
\affiliation{%
  \institution{The Charles Stark Draper Laboratory, Inc.}
  \city{Cambridge}
  \state{Massachusetts}
  \country{USA}
}
\author{Lili Su}
\affiliation{%
  \institution{Northeastern University}
  \city{Boston}
  \state{Massachusetts}
  \country{USA}
}

\renewcommand{\shortauthors}{Mclaughlin et al.}

\begin{abstract}
Machine learning models often need to adapt to new data after deployment due to structured or unstructured real-world dynamics. 
The Continual Learning (CL) framework enables continuous model adaptation, but most existing approaches either assume each task contains sufficiently many data samples or that the learning tasks are non-overlapping.    
In this paper, we address the more general setting where each task may have a limited dataset, and tasks may overlap in an arbitrary manner without a priori knowledge. 
This general setting is substantially more challenging for two reasons. On the one hand, data scarcity necessitates effective contextualization of general knowledge and efficient knowledge transfer across tasks. 
On the other hand, unstructured task overlapping can easily result in negative knowledge transfer. 
To address the above challenges, we propose an adaptive mixture-of-experts (MoE) framework 
over pre-trained models that progressively establishes similarity awareness among tasks. 
Our design contains two innovative algorithmic components: incremental global pooling and instance-wise prompt masking. 
The former mitigates prompt association noise through gradual prompt introduction over time. 
The latter decomposes incoming task samples into those aligning with current prompts (in-distribution) and those requiring new prompts (out-of-distribution).
Together, our design strategically leverages potential task overlaps while actively preventing negative mutual interference in the presence of per-task data scarcity.  
Experiments across varying data volumes and inter-task similarity show that our method enhances sample efficiency and is broadly applicable.

\end{abstract}

\maketitle

\section{Introduction}
\label{sec: intro}

\begin{figure}[b]
    \centering
    \includegraphics[width=0.92\linewidth]{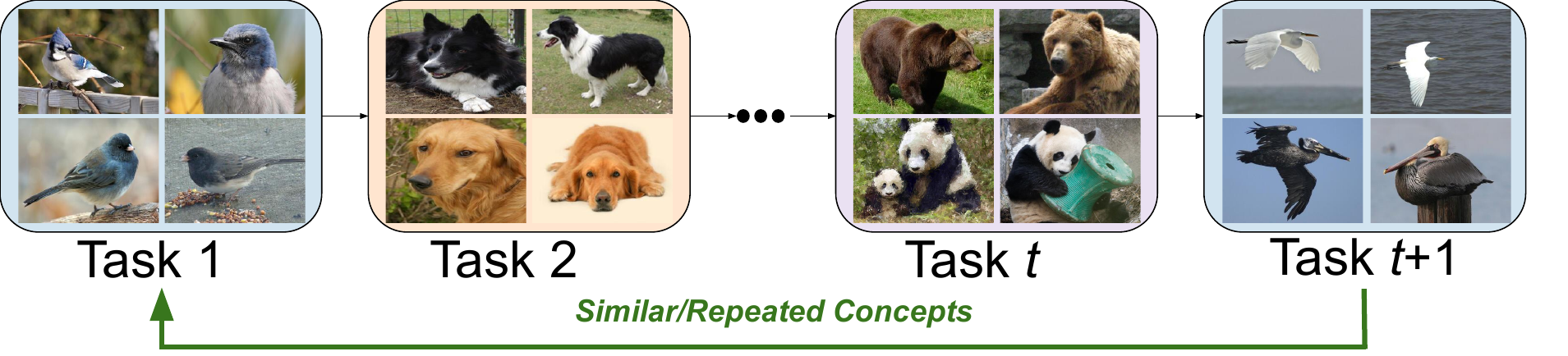}
    \vskip -0.5\baselineskip 
    \caption{ \footnotesize As tasks of interest evolve over time, the learner may encounter similar or recurring concepts. For example, Task 1 and Task $t+1$ 
    may both involve the high-level category of birds.
    }
    \label{fig: overlap tasks}
\end{figure}

Machine learning models often need to adapt to new data after deployment due to structured or unstructured dynamics in learning tasks. 
The na\"ive solution of retraining from scratch with all previously encountered data is prohibitively -- and often unnecessarily -- costly, as it ignores both the knowledge gained from historical data and the correlations among tasks. 
More importantly, such na\"ive solutions are often infeasible due to privacy constraints and data regulations. 
The Continual Learning (CL) framework enables continuous model adaptation. Yet, most existing approaches either assume each task contains sufficiently many data samples or that the learning tasks are non-overlapping \citep{wang2024comprehensive,mccloskey1989catastrophic}. 
In many modern real-world applications, the amount of data available per task can be scarce, and the tasks themselves are often overlapping. 
For instance, in fraud detection, each fraud type may be represented by a few cases (data scarcity), but fraudulent patterns often share features across transaction types (task overlap).  
Similarly, in image classifications, tasks of interests evolve over times but may share similar or recurring task concepts, as illustrated in Fig.\ref{fig: overlap tasks}.  

In this paper, we make an attempt at addressing data scarcity and task overlap simultaneously. For the sake of fair comparison with existing CL benchmarks, we focus on classification tasks, while extending the method to regression problems is left for future work. It is well-known that traditional CL methods often suffer from the issue of catastrophic forgetting -- the tendency of a model to lose previously acquired knowledge on seen tasks when trained on new tasks \citep{mccloskey1989catastrophic}.  
Our general setting is substantially more challenging than traditional settings for two reasons. Data scarcity necessitates effective contextualization of general knowledge and efficient knowledge transfer across tasks, which is already highly nontrivial. 
On top of that, unstructured task overlap—while providing a fundamental basis for information sharing—can also easily lead to negative knowledge transfer. 
Thus, we believe this more general setting calls for a fundamental rethinking of CL method designs.  

To address per-task data scarcity and to make effective use of existing computation, we pursue knowledge transfer along two complementary dimensions: 
\begin{itemize}
\item[(1)] knowledge transfer from a pre-trained foundation model via prompting, 
\item[(2)] inter-task knowledge transfer via adaptively establishing and strengthening a pool of prompts (i.e., mixture-of-experts [MoE]).  
\end{itemize} 
Intuitively, the former allows us to extract task-relevant knowledge from a well-trained generalist  by training a compatible domain expert (i.e., the prompt), whereas the latter iteratively enhances the expertise of the experts by collaboratively refining the prompts based on newly observed data.  

Harnessing the capabilities of pre-trained Vision Transformers (ViTs), recent advancements in rehearsal-free CL have pivoted towards prompt-based strategies \citep{wang2022learning, wang2022dualprompt, wang2022s, smith2022coda}. By learning only a small set of prompt parameters that are injected into a frozen ViT, these approaches substantially reduce the number of trainable parameters and, consequently, the sample complexity required for adapting to new tasks.  
A prominent direction in prompt-based CL is the expert-prompt strategy \citep{wang2022dualprompt, wang2022s}, where each task is assigned its own prompt parameters trained solely on task-specific data. 
While effective, this design implicitly assumes the availability of sufficient task-specific supervision and clear task delineation. 
An alternative is the global prompt pool \cite{wang2022learning, smith2022coda}, which allows prompts to be shared and updated across all tasks. While this encourages reuse, it introduces optimization noise due to routing errors.  Routing errors create a fundamental dilemma in data-scarce regimes. As local data become more limited, effective learning increasingly relies on knowledge transfer from other tasks or pre-trained models. However, data scarcity also amplifies routing noise, which can in turn lead to negative knowledge transfer. 
An illustration on this can be found in Fig.\,\ref{fig:incrementalpool} Section \ref{sec: Methods}. 

\begin{figure*}[!htbp]
\centering
\includegraphics[width=.98\linewidth, height=0.25\linewidth]{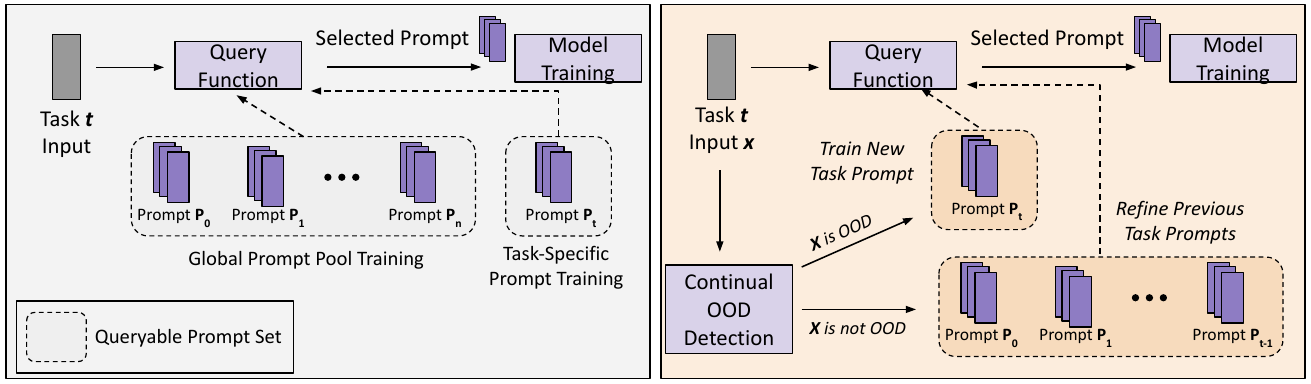} 
\caption{\small Comparison of Prompt Routing Strategies. (Left) When training on task $t$, typical methods select prompt models in one of two ways: either by querying an unstructured global pool or by rigidly training a prompt designated for task $t$. (Right) We employ an out-of-distribution (OOD) detector to dynamically determine which prompts may be queried. Inputs detected as novel (OOD) are routed to a task-specific prompt to learn new concepts, whereas for familiar samples, we query the set of previous task prompts to facilitate knowledge reuse.
}
\label{fig:method_overview}
\end{figure*}

\vspace{0.5em}
\noindent {\bf Contributions.}
Our contributions can be summarized as follows. 
\begin{itemize}
\item We identify a general, yet critically overlooked, setting in modern CL practice that necessitates a fundamental rethinking of cross-task knowledge sharing.   
\item  To enable effective knowledge sharing despite data-scarcity, we challenge the rigid routing structures of prior prompt-based approaches and propose an adaptive and similarity-aware routing mechanism. 
Our design contains two innovative algorithmic components: incremental global pooling and instance-wise prompt masking. 
Together, they effectively leverage potential task overlaps while preventing negative mutual interference in the presence of per-task data scarcity.  

\item Our method establishes a connection between CL and out-of-distribution detection \citep{kim2022continual, kim2025open}: by measuring the similarity between incoming data and previously encountered tasks, the model can more effectively allocate prompts to separate task heterogeneity and high sampling noise.  
We hope that this connection can inspire the design of future CL methods by drawing on tools and insights developed in the out-of-distribution detection community. 
\item Experiments across varying data volumes and inter-task similarity show that our method enhances sample efficiency and is broadly applicable.
\end{itemize}

\section{Related Work}
The primary objective of CL is to enable machine learning models to acquire new skills over time without erasing previously acquired knowledge, i.e., \textit{catastrophic forgetting}.  CL has received intensive research attention due to its broad applications. Several recent surveys provide comprehensive overviews of the foundations, challenges, and recent advances in CL \citep{wang2024comprehensive,de2021continual,shi2025continual}. 
In the following, we briefly touch on a few lines of work that are closely related to our setting.
Interested readers are referred to the corresponding surveys for further details.


\vspace{0.2em}
\noindent \textit{Regularization-based methods} aim to preserve knowledge by constraining parameter updates. Approaches such as Elastic Weight Consolidation \citep{kirkpatrick2017overcoming} and Synaptic Intelligence \citep{zenke2017continual} introduced penalty terms to the loss function, protecting weights deemed critical for previous tasks. Knowledge distillation \citep{li2017learning} is a frequent companion to these methods, transferring output probabilities from an old model to the current one to stabilize training \citep{aljundi2018memory}. However, these methods can struggle in settings where the task identity is unavailable during inference.

\vspace{0.2em}
\noindent \textit{Parameter isolation methods} \citep{rusu2016progressive, mallya2018packnet, serra2018overcoming} mitigated interference by allocating different subsets of model parameters to specific tasks. Techniques such as supermasks \citep{wortsman2020supermasks} or pruning \citep{mallya2018piggyback} ensure that new tasks do not disrupt existing pathways. Although effective in removing forgetting, these architectures often scale poorly, as the model complexity may grow linearly with the number of tasks.

\vspace{0.2em}
\noindent \textit{Rehearsal-based/rehearsal-free methods.}  Rehearsal-based methods \citep{rebuffi2017icarl, lopez2017gradient, chaudhry2019efficient, buzzega2020dark,verwimp2021rehearsal, yoon2021online, zhang2022simple} generally yield the strongest performance in class incremental learning settings. These methods maintain a memory buffer of samples from past distributions (experience replay) or generate them via generative models (pseudo-rehearsal). Despite their efficacy, the storage of raw data raises significant concerns about privacy. 
As a result, the field of \textit{rehearsal-free} methods \citep{wang2022dualprompt, smith2022coda} has gained significant attention, which is the topic of this work.

\vspace{0.2em}
\noindent \textit{Prompt-based CL} has emerged as a compelling parameter-efficient alternative to traditional fine-tuning. Inspired by advances in Natural Language Processing (NLP), these methods freeze a pre-trained backbone (e.g., a ViT) and append a small set of learnable vectors (prompts) to the input sequence. 
The core challenge in this domain is designing mechanisms to select and update prompts dynamically. L2P \citep{wang2022learning} introduced the concept of a shared prompt pool, utilizing a key-query mechanism to retrieve relevant prompts for a given input. DualPrompt \citep{wang2022dualprompt} refined this by architecturally separating prompts into task-agnostic and task-specific components to better balance stability and plasticity. Other variants, such as S-Prompts \citep{wang2022s}, focused purely on task-specific learning, while CODA-Prompt \citep{smith2022coda} employed an attention-based weighted summation to construct prompts from a pool. Consistent Prompting \citep{zhao2024consistent} further emphasizes the stability of prompt behavior across distributions.

\section{Problem Formulation}
\label{sec: formulation}
We study the problem of rehearsal-free CL, 
where a classification model $\calM$ is trained sequentially on a series of tasks without storing past data. Formally, the learner observes a sequence of $T$ datasets \(\mathfrak{S} = \{\calS_0, \calS_1, \dots, \calS_{T-1}\} \), with each $\calS_t = \{(x^i_t, y^i_t)\}_{i=1}^{n_t}$ where $x^i_t\in \calX$ is the input feature vector, $y_t^i\in \calC$ is a label. 
Let $\calC_t = \{y_t^i: i=1, \cdots, n_t\}$ denote the collection of distinct labels of task $\calS_t$.  
As mentioned in Section \ref{sec: intro}, we consider the general setting wherein $n_t$ may be small, and $\calC_t\cap \calC_{t^{\prime}} \not=\emptyset$ for some $t^{\prime} \not=t$.

To avoid overfitting and to better capture the generalization capability of the trained model, we measure its performance in terms of the average generalization error of the tasks in $T$. Formally, let $d_t$ denote the underlying distribution for which $(x_t^i, y_t^i)\overset{\iid}{\sim} d_t$. 
We would like to obtain a model $\calM$ that minimizes the average generation error  
\begin{align}
\label{eq: CL goal}
\min_{\calM} \frac{1}{T}\sum_{t=1}^T \mathbb{E}_{(x,y)\sim d_t}[\indc{\calM(x) \not= y}], 
\end{align}
where $\calM(x)$ denotes the label prediction, and $\indc{\calM(x) \not= y}$ is an indicator function which takes value 1 if $\calM(x) \not= y$, and 0 otherwise.  
At the inference time, the trained model is not told to which task the input feature encountered $x$ belongs. 
We focus on rehearsal-free methods, i.e., no subset of samples from previous tasks is stored. 

\section{Preliminaries: Prompt-Based CL}
We introduce the relevant backgrounds and notations for the prompt-based continual learning and mixture-of-experts that our method builds upon. 

\vspace{0.3em}
\noindent {\bf Model Structure.}
In the prompt-based CL, model to be learned is decomposed into three components: 
(1) a pre-trained backbone foundation model $f_{\theta}$ (e.g.\, a pre-trained ViT), which encodes the general knowledge obtained in the pre-training; 
(2) a collection of prompts $\calP$ that can be flexibly combined with $f_{\theta}$ to extract knowledge that are specific to the given task sequence $\mathfrak{S}$, and (3) classification head $g_\psi$, where $\psi$ denotes learnable parameters \citep{dosovitskiy2020image,wang2022learning}. 

For any given $x$, its label prediction is determined as 
\begin{align}
\label{eq: label prediction}
\hat{y}(x) = g_{\psi}\circ \pth{f_{\theta}(x) \oplus p(x)}, 
\end{align}
where $p(x)$ is a set of prompts that are relevant to input $x$, and, with slight abuse of notation, $\oplus$ represents the chosen modulation operation between the prompts and the pre-trained backbone. 

Throughout the CL process, the backbone model $f_{\theta}$ remains fixed, whereas $\psi$ is repeatedly updated for each task. In addition, for any given task, the prompts are updated according to some rules of determining $p(x)$, which we will elaborate on soon.

\vspace{0.3em} 
\noindent {\bf ViT as the Pre-trained Backbone $f_{\theta}$.}
An input $x\in\calX:=\reals^{H\times W\times C}$ is split into $N=HW/P^2$ patches, each linearly projected to $d$-dimensional tokens. Together with a class token $e_{\text{cls}}$ and positional embeddings, the input to the first layer is
\begin{equation}
h^0 = [e_{\text{cls}}; e_p^1;\dots;e_p^N] + e_{\text{pos}} 
   \in \reals^{L^h\times d},
\end{equation}
where $L^h=N+1$ is the sequence length.  

A transformer encoder layer applies multi-head self-attention (MSA) followed by an MLP. For the $l$-th layer, with input $h^l\in\reals^{L^h\times d}$,
\begin{align}
&\text{MSA}(h^Q,h^K,h^V) = \text{Concat}(h^1,\dots,h^m)W^O, \\
& h^i = \text{Attention}(h^Q W^Q_i, h^K W^K_i, h^V W^V_i)
\end{align}
where $W^O,W^Q_i,W^K_i,W^V_i$ are learned projections, $m$ is the number of heads, and in vanilla ViT we set $h^Q=h^K=h^V=h^l$.

\vspace{0.3em}
\noindent {\bf Prompt Structure.} 
We employ a Vision Transformer (ViT) backbone $f_{\theta}$, modulated by a set of learnable prompts $\calP$, where each $p\in \calP$ is a parameter vector $p \in \reals^{L_p \times d}$. 
Our framework supports common modulation strategies such as prompt tuning \cite{lester2021power}, where prompts are prepended to the input sequence, and prefix tuning \cite{li2021prefix}, where they are prepended to keys and values in MSA layers.
The choice of modulation and layers is treated as a hyperparameter.

\noindent {\bf Query-Based Selection.} 
To handle non-stationary data distributions, prompts are typically selected dynamically based on input relevance. 
We focus on instance-wise query mechanisms \cite{wang2022learning, wang2022dualprompt}, where a query function $q(\cdot, \cdot)$ measures the similarity between an input sample $x$ and a prompt's key $\phi(p)$.
Commonly, the query operates in the pre-trained embedding space (e.g., using the [CLS] token $z(x) = f_\theta(x)[0]$), and $\phi(p)$ is a learnable key vector associated with prompt $p$.
The relevance score is defined via cosine similarity:
\begin{align}
q(x, \phi(p)) : = \frac{\langle z(x), \phi(p) \rangle}{\|z(x)\| \|\phi(p)\|}.
\end{align}
For any input $x$, we select a subset of prompts $p(x)$ by retrieving the top-$K$ matches from the available pool $\calP$:
\begin{align}
\label{eq: selection_mechanism}
p(x) : = \argmax_{\substack{\calR \subseteq \calP, |\calR| = K}} \sum_{p \in \calR} q(x, \phi(p)).
\end{align}
While soft aggregation \cite{smith2022coda} is an alternative, we focus on hard selection for clarity.

\noindent {\bf Optimization.} 
Let $\calL$ denote the prediction loss (e.g., cross-entropy). The prediction is given by $\hat{y}(x) = g_\psi(f_\theta(x) \oplus p(x))$, where $\oplus$ denotes the prompt modulation.
The prompts and the classification head $g_\psi$ are updated by minimizing the empirical risk over the current task samples $\calS_t$:
\begin{align}
\label{eq: optimization_objective}
\min_{p\in \calP, ~ \psi} \frac{1}{n_t} \sum_{i=1}^{n_t} \mathcal{L}\!\left(y_i, \, \hat{y}(x_i)\right).
\end{align}
Intuitively, the more frequently a prompt $p$ is selected via Eq. \ref{eq: selection_mechanism}, the more it contributes to—and is updated by—the task loss.

\noindent {\bf Existing Strategies.} 
Existing prompt-based CL methods generally fall into two categories based on how they manage the pool $\calP$:
\begin{itemize}
    \item \textbf{Global Prompt Pool} \cite{wang2022learning, smith2022coda}: The pool size $|\calP| = N_p$ is fixed. Prompts are initialized once (randomly or via heuristics) and shared across all tasks. During training, the entire pool is available for selection, and updates are driven purely by the query mechanism defined above.
    \item \textbf{Task-Specific Prompts} \cite{wang2022dualprompt, zhang2022simple}: A new, disjoint set of prompts $\calP_t$ is created for each task $t$, such that $\calP_t \cap \calP_{t'} = \emptyset$ for $t \neq t'$. During training on task $t$, the search space in Eq. \ref{eq: selection_mechanism} is restricted to only the newly created prompts $\calP_t$, explicitly isolating tasks to prevent interference.
\end{itemize}

\section{Methodology}
\label{sec: Methods}

In this work, we propose a novel adaptive routing mechanism designed to facilitate effective knowledge transfer in the presence of per-task data scarcity. 
Specifically, we introduce two innovative and complementary algorithmic components: (A) \textbf{incremental global pooling} and (B)  \textbf{instance-wise prompt masking}. 
At a high-level, component (A) is used to control the routing errors in somewhat early stage of the CL procedure. 
In component (B), via a novel masking strategy, we essentially decompose the current task $\calS_t$ into samples that can be well-explained by existing prompts and the ones that are nearly ``orthogonal''.

\subsection{Incremental Global Pooling}
We maintain a global prompt pool that grows incrementally as we encounter more and more tasks. 

Let $\tilde{\calP}_t$ denote the incrementally constructed pool up to task $t$, such that $|\tilde{\calP}_t| \le |\tilde{\calP}_{t'}|$ for any $t' > t$. While the prompt indices of $\tilde{\calP}_t$ form a subset of $\tilde{\calP}_{t'}$, the prompts themselves are continuously updated, so strictly speaking $\tilde{\calP}_t \not\subseteq \tilde{\calP}_{t'}$. 
At the start of task $t$, new prompts are incorporated, i.e., $\tilde{\calP}_t = \tilde{\calP}_{t-1}^{\prime}\cup \{\text{new prompts}\}$, where $\tilde{\calP}_{t-1}^{\prime}$ is the collection of updated prompts at the end $(t-1)$-th task training. 
We initialize these new prompts to be orthogonal to the current prompt set $\tilde{\calP}_{t-1}^{\prime}$, following recent practice \cite{smith2022coda}. This initialization minimizes the initial overlap between the new prompt training and past knowledge.
All prompts in $\tilde{\calP}_t$ are available for routing the samples in $\calS_t$, and may possibly be updated during training.

\begin{figure}
    \centering
    \includegraphics[width=\linewidth]{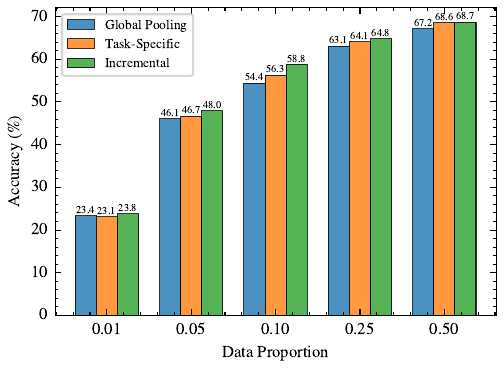}
    \caption{Final Average Accuracy on ImageNet-R across varying training set sizes (expressed as a fraction of the full dataset). While global pooling enables effective transfer in scarce-data regimes, it suffers from interference as data availability increases. Incremental expansion mitigates this by reducing routing errors in early learning stage, simultaneously enabling the refinement of consolidated knowledge and the allocation of fresh capacity for novel knowledge. 
    }
    \label{fig:incrementalpool}
\end{figure}

It is worth noting that, for each task $t$, 
this algorithmic component can already substantially constrain negative interference across tasks, especially when per-task data are scarce. We study this phenomenon in Figure \ref{fig:incrementalpool}, by training an identical set of prompt parameters on ImageNet-R 
with various prompt selection methods using varying proportions of the training set. 
Despite knowledge sharing becomes a necessity when isolated task prompts lack sufficient samples, only in the extreme low-data regime (e.g., $1\%$ of training data), global pooling has superior performance to task-specific learning, indicating that existing global pooling methods do not lead to successful knowledge sharing. 
The superior performance of our approach can be attributed to the following rationale.  
When $t$ is small, only a limited number are updated to encode task sequence relevant information in the prompt pool, leaving the majority in their original state.
When per-task data is scarce, even the updated prompts retain significant noise, blurring the distinction between them and the unmodified prompts. By restricting the pool size when $t$ is small, we increase the selection frequency of previous updated prompts, enabling the refinement of consolidated knowledge. By continuously introducing new prompts, fresh capacity is allocated for the novel knowledge contained in later tasks.

\subsection{Instance-Wise Prompt Masking.} 
\label{subsec: instance-wise mask}
Task overlap is both an enabler and a threat: it enables knowledge transfer yet, in the presence of unknown relatedness and dataset noise, it inherently risks catastrophic negative interference. 
We introduce a novel masking strategy that acts on each sample in the current tasks $\calS_t$. 

Fix any arbitrary task $t$. 
Let $m(x, \calH_{t-1}) \in \{0, 1\}$ denote a masking function that determines which subgroup of $\tilde{\calP}_{t}$ one should route a input covariate $x$ of task $\calS_t$ to. 
Here, $\calH_{t-1}$ denotes lightweight task summary (e.g., embedding statistics, feature covariances, or uncertainty measures) up to the end of the $(t-1)$-th task.
If $m(x, \calH_{t-1}) =0$, we mask away the new prompts created, restricting the query-based prompt selection for sample $x$ to $\tilde{P}_{t-1}^{\prime}$; that is, we restrict the optimization in  Eq.\,\eqref{eq: selection_mechanism} from over $\calP$ to over $\tilde{P}_{t-1}^{\prime}$.  
If $m(x, \calH_{t-1}) =1$, we we mask away existing prompts, restricting the query-based prompt selection for sample $x$ to new prompts. Let $\calP_t(x)$ denote the prompt pool for $x$.  
\begin{align}
\calP_t(x) =
\begin{cases}
\tilde{\calP}_{t-1}^{\prime}, & \text{if } m(x, \calH_{t-1}) =0  \\
\{\text{new prompts}\}, & \text{if } m(x, \calH_{t-1}) =1.  
\end{cases}
\end{align}
Importantly, this masking strategy does not replace the underlying query mechanism (e.g., learnable keys), but rather \textit{filters} the search space available to it.

Intuitively, this instance-wise prompt can help in decomposing the current task $\calS_t$ into samples that can be well-explained by existing prompts and the ones that are nearly ``orthogonal''. The ``orthogonality'' may stem from two sources: fundamental statistical heterogeneity (i.e., distributional differences) and the random sampling noise. The former persists as $t$ increases, while the latter diminishes as more relevant samples are observed over time.

\subsection{Masking Function Design}

The effectiveness of the instance-wise prompt masking in Section \ref{subsec: instance-wise mask} depends crucially on the design of the masking functions, which is the focus on this subsection.  
A poor design of $m$ can yield inferior overall performance compared to using incremental pooling alone.

The concept of history-auxiliary sample decomposition mirrors the central challenge of out-of-distribution (OOD) detection: samples well-explained by existing prompts are treated as in-distribution (ID), while those that are not are considered OOD.
To implement the masking function $m$, we adopt the Relative Mahalanobis Distance (RMD) \citep{ren2021simplefixmahalanobisdistance} as the discriminative metric. 
\begin{definition}
\label{def: RMD}
Let $f: \calX \to \reals^d$ be a given feature extractor. Let $\calD =\{(x^i, y^i)\}_{i=1}^n$ be a given classification dataset with label set $\calC$. 
For each $c\in \calC$, define the class-wise feature mean as $\mu_c : = \frac{1}{|\{i\in [n]: ~ y^i=c\}|} \sum_{i=1}^n f(x^i)\indc{y^i = c}$. 
Let $\Sigma$ denote the overall feature variance matrix. 
Suppose that $\Sigma$ is of full rank. 
The Mahalanobis distance of a given input $x$ to class $c$ is defined as 
\[
\text{MD}_c^f(x) = \pth{f(x) - \mu_c} \Sigma^{-1} \pth{f(x) - \mu_c}. 
\]
Furthermore, let $\widebar{\text{MD}}_c^f = \frac{1}{|\{i\in [n]: ~ y^i=c\}|} \sum_{i=1}^n \text{MD}_c^f(x^i) \indc{y^i = c}$. 
The Relative Mahalanobis Distance of a given $x$ to class $c$ is defined as 
\[
\text{RMD}^f(x) = \min_{c\in \calC} \pth{\text{MD}_c^f(x) - \widebar{\text{MD}}_c^f}.  
\] 
\end{definition}
We select RMD primarily for its state-of-the-art performance in \textit{near-OOD} detection. We prioritize the near-OOD setting as it addresses the critical challenge where inputs may be perceptually similar to previous samples (potentially resulting in high query similarity), but semantically distinct \cite{mukhoti2023raising}. This assumption is regarded as more practical for realistic open-world applications compared to \textit{far}-OOD detection (e.g., noise vs. images) \cite{fort2021exploring}. Furthermore, RMD is highly efficient to apply within prompt-based CL, as it leverages the pre-computed backbone features. We discuss the specific implementation and computational details below. 

\vspace{0.3em} 
\noindent \textbf{Feature Statistics.}  
In our rehearsal-free CL, dataset $\calD$ is not given all at once, and cannot be stored as along the task sequence $\mathfrak{S}$.  
To obtain $\mu_c$ and $\Sigma$ under this data access constraint, we maintain task summary statistics $\calH_{t-1} = \{\calH_{t-1}^{\text{class}}, \calH_{t-1}^{\text{global}}\}$ for all tasks up to $t$, estimated in the feature space of the pre-trained model $f_{\theta}$.  
\begin{itemize}
    \item \textit{Per-class statistics:}  
    $\calH_{t-1}^{\text{class}} = \{(\mu_c^{t-1}, \Sigma_c^{t-1}, n_c^{t-1}, \widehat{\text{MD}}_c^{f_{\theta}, t-1}) : c \in \cup_{r=0}^{t-1}\calC_{r}\}$,  
    where $\mu_c^{t-1} \in \reals^d$ and $\Sigma_c^{t-1} \in \reals^{d\times d}$ denote the empirical mean and covariance of the extracted features for class $c$, $n_c$ is the total number of samples with label $c$ in all previous tasks, and 
    $\widehat{\text{MD}}_c^{f_{\theta}, t-1}$ is an approximation of $\widebar{\text{MD}}_c^{f_{\theta},t-1}$.  
    \item \textit{Global background statistics:}  
    $\calH_{t-1}^{\text{global}} = (\mu_{\text{bg}, t-1}, \Sigma_{\text{bg}, t-1})$, computed overall features observed so far.
\end{itemize}

At the end of task $t$, the per-class statistics $(\mu_c, \Sigma_c, n_c)$ and global statistics $(\mu_{\text{bg}}, \Sigma_{\text{bg}})$ are updated using extracted features from $\calS_t$ via standard online mean and covariance updates. 
The approximation $\widehat{\text{MD}}_c^{f_{\theta}}$ is updated as 
\begin{align*}
\widehat{\text{MD}}_c^{f_{\theta}, t} &= \frac{n_c^{t-1}}{n_c^{t-1} + \Delta n_c^t} \widehat{\text{MD}}_c^{f_{\theta}, t-1}  \\
& \quad \quad + \frac{1}{n_c^{t-1} + \Delta n_c^t} \sum_{i=1}^{n_t}     \pth{f_{\theta}(x_t^i) - \mu^t_c} \Sigma_{\text{bg}, t}^{-1} \pth{f_{\theta}(x_t^i) - \mu^t_c}, 
\end{align*}
where $\Delta n_c^t = \sum_{i=1}^{n_t} \indc{y_t^i = c}$. In practice, the presence of noise in real-world data generally ensures that the covariance matrices $\Sigma_c$ and $\Sigma_{\text{bg}}$ are full rank. When this assumption is violated, we apply a standard regularization technique by adding $\epsilon \bf{I}$ with a small $\epsilon$.

\vspace{0.3em}

\noindent \textbf{Masking Function.}  

Let 
\[
\text{RMD}_t(x) := \min_{c\in \cup_{r=0}^{t-1}\calC_r}\pth{\text{MD}_c^{f_{\theta}, t-1}(x) - \widehat{\text{MD}}_c^{f_{\theta}, t-1}},
\]
where $\text{MD}_c^{f_{\theta}, t-1}(x) = \pth{f_{\theta}(x) - \mu^{t-1}_c} \Sigma_{\text{bg}, t-1}^{-1} \pth{f_{\theta}(x) - \mu^{t-1}_c}.$ 
We propose the following masking function: 
\begin{align*}
m(x, \calH_{t-1}^{\text{class}}, \calH_{t-1}^{\text{global}}) =
\begin{cases}
 0, & \quad \text{if} ~ \text{RMD}_t(x) \le \tau_t\\
 1, & \quad \text{if} ~~ \text{RMD}_t(x) >\tau_t, 
\end{cases}
\end{align*}
where 

the task-specific decision threshold $\tau_t$ is determined as the empirical $q$-th quantile of this buffer:
\begin{align}
\tau_t = Q_q\!\left(
\{\text{RMD}_{t'}(x) : x \in \calD_{t'},\; t' \le t\}
\right),
\end{align}
where $Q_q(\cdot)$ denotes the empirical $q$-th quantile (e.g., $q = 0.99$) of the given set.

We empirically validate our choice of masking function compared with existing routing methods in Table \ref{table:query_auc}, where we analyze the prompt retrieval scores for seen versus unseen task samples throughout training on ImageNet-R. To quantify the reliability of detecting genuine task shifts, we report the Area Under the Curve (AUC) metric, which measures the distributional separation between the similarity scores of known tasks and those of disjoint, unseen tasks (i.e., future classes of ImageNet-R). Ideally, unseen tasks should yield distinctively lower scores, but standard query mechanisms exhibit poor separation, with AUC dropping as low as 0.6 in some settings. This indicates that similarity scores alone may be unreliable for routing samples in a global prompt pool.

{\setlength\intextsep{0pt}
\begin{table}[!ht]
\begin{tabular}{@{}ccc@{}}
\toprule
              & ImageNet-R & 5-Dataset \\ \midrule
Learnable Key (\cite{wang2022learning, wang2022dualprompt}) & .602       & .749      \\
Task Centroids (\cite{wang2022s})     & .633       & .698      \\
Task Classifier (\cite{le2025mixtureexpertsmeetspromptbased})    & .749       & .791      \\
Online RMD (Ours)    & .877       & .819      \\ \bottomrule
\end{tabular}
\caption{
Task Separability Analysis. Average AUC scores for discriminating between seen and unseen tasks. We compare our proposed method against popular query score mechanisms (Learnable Keys, Task Centroids, Task Classifiers) on the ImageNet-R and 5-Dataset benchmarks.
}
\label{table:query_auc}
\end{table} 
}

\section{Experiments}
\label{sec: experiments}

\subsection{Evaluation Setup}

\textbf{Benchmark:}  We evaluate our method on several standard and challenging continual learning benchmarks. Split CIFAR-100 \cite{krizhevsky2009learning_cifar} consists of 100 classes partitioned into 10 incremental tasks. Split ImageNet-R \cite{ridnik2021imagenet21k} contains 200 classes featuring diverse artistic renditions, split into 10 tasks to test robustness against significant intra-class style variations. 5-Datasets \cite{ebrahimi2020adversarial} provides a sequence of five distinct datasets: CIFAR-10, MNIST, Fashion-MNIST, SVHN, and notMNIST, to measure performance under extreme domain shifts between tasks. As another benchmark of performance under domain shift, we include DomainNet \cite{peng2019moment}, which introduces visually distinct domains (e.g., Clipart, Quickdraw) sequentially. Finally, we adopt the mixed-dataset protocol from \citet{ke2021adapting}, which we denoted as M-EMNIST and M-CifarCeleb. These benchmarks present a sequence of both similar and dissimilar tasks, where task similarity is emulated by sampling data from the same ``client" in datasets commonly used for federated learning. For both mixed datasets, we evaluate on a 20-task setting, with 10 similar and dissimilar tasks respectively.

\textbf{Compared Methods.} Our evaluation compares with five state-of-the-art prompting methods: L2P \cite{wang2022learning}, DualPrompt \cite{wang2022dualprompt}, S-Prompt \cite{wang2022s}, CODA-Prompt \cite{smith2022coda}, and HiDe-Prompt \cite{wang2023hierarchical}. Following \cite{wang2023hierarchical}, we modernize the S-Prompt and use a shared classification head, denoting the resulting method as S-Prompt++. We report Final Average Accuracy (FAA), Cumulative Average Accuracy (CAA), and Final Forgetting Measure (FFM). 

\textbf{Implementation:}
Our proposed routing policy is agnostic to the underlying prompt structure (e.g., prefix-tuning vs. prompt-tuning), similarity scoring function, and auxiliary strategies beyond prompt selection while training. In our experiments, we instantiate our method using two distinct architectures: we adopt HiDe-Prompt \cite{wang2023hierarchical} as our primary base recipe for main result comparisons, and utilize S-Prompt++ \cite{wang2022s} as a simplified baseline for ablation studies (Figures \ref{fig:incrementalpool}, \ref{fig:prompt_usage}, \ref{fig:ablation}).

Unless otherwise specified, we follow standard protocols consistent with previous work \cite{wang2023hierarchical, smith2022coda}. We employ an ImageNet-21k  pre-trained ViT-B \cite{ridnik2021imagenet} backbone optimized via Adam with a batch size of 128. We set the prompt length $L_p=5$ and quantile $q=0.95$ when determining the OOD threshold. All input images are resized to 224×224 and normalized to [0,1].

\subsection{Results and Analysis}
\begin{figure}[th]
    \centering
    \includegraphics[width=\linewidth, trim={0 0.3cm 0 0}]{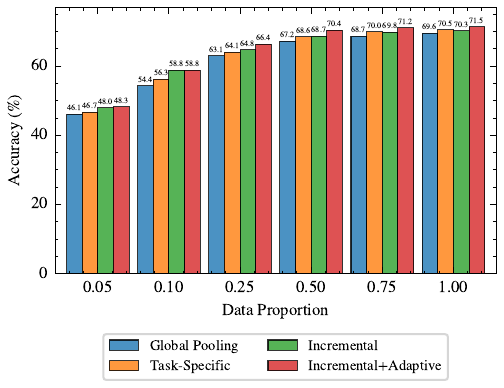}
    \caption{Impact of training data availability on routing method performance on ImageNet-R.}
    \label{fig:ablation}
\end{figure}

\textbf{Comparative Analysis of Prompt Routing Policies.} We begin by evaluating our dynamic routing policy against the two dominant paradigms in the literature: global pooling and task-specific allocation. Figure ~\ref{fig:ablation} illustrates the performance of these strategies across varying training data proportions. 

We first evaluate the proposed dynamic routing policy against the two predominant paradigms in prompt-based continual learning: global pooling and task-specific allocation. Figure~\ref{fig:ablation} illustrates the comparative performance of these strategies across varying training data proportions.
Our analysis reveals that the proposed method is particularly effective in data-constrained regimes, where the choice of routing policy determines the efficiency of knowledge acquisition. 
In contrast, task-specific methods exhibit limited scalability in these settings; because parameter sets are strictly disjoint, the model is unable to exploit cross-task redundancies, necessitating a high volume of samples to optimize each new prompt.
Conversely, while global-pool methods facilitate parameter sharing, the lack of structured routing results in a high degree of inter-task interference and representational collapse. 
In the low-data regime, this interference frequently outweighs the benefits of sharing, causing global-pool methods to underperform relative to isolated baselines.
The proposed adaptive routing policy circumvents this trade-off by dynamically modulating the degree of parameter sharing based on the input distribution.
By selectively retrieving relevant features while isolating task-specific refinements, our method maintains the high plasticity required for data-efficient learning without the performance degradation associated with unstructured sharing. 
Consequently, we observe consistent improvements over baseline routing methods, with the most significant margins occurring when training data is most scarce.

\textbf{Knowledge Accumulation and Reuse in Extended Task Streams.}
We utilize the 5-Datasets and DomainNet benchmarks (Tables~\ref{table:5datasets_scaling} and \ref{table:domainnet_scaling}) to evaluate model scalability over extended horizons. These datasets are representative of practical settings where the model encounters a diverse range of inputs, e.g., new classes from different datasets (5-Datasets),  or shared classes across distinct visual domains (DomainNet).

\begin{table}[h]
    \centering
    \caption{Performance on 5-Datasets. We present FAA ($\uparrow$) across different task scales (5, 10, and 15 tasks).} 
    \renewcommand\arraystretch{1.2}
    \setlength\tabcolsep{4pt}
    \small{
    \resizebox{0.38\textwidth}{!}{ 
    \begin{tabular}{l|ccc}
     \hline
      \multirow{2}{*}{Method} & \multicolumn{3}{c}{5-Datasets} \\
        & 5-task & 10-task & 15-task \\
        \hline
       L2P & 81.84 & 78.45 & 75.22 \\ 
       DualPrompt & 77.91 & 74.33 & 71.15 \\ 
       S-Prompt++ & 86.06 & 82.71 & 79.58 \\ 
       CODA-Prompt & 64.18 & 61.52 & 58.83 \\ 
       HiDe-Prompt & 93.83 & 90.25 & 87.12 \\ 
       Ours & \textbf{93.92}  & \textbf{91.18} & \textbf{88.45} \\
       \hline
    \end{tabular}
    } 
}
    \label{table:5datasets_scaling}
\end{table}

As shown in Table~\ref{table:5datasets_scaling}, our method achieves the highest Final Average Accuracy (FAA) across all task settings. While all methods experience a performance decline as the sequence length increases, the margin of our improvement grows with task length.

\begin{table}[h]
    \centering
    \caption{Performance on DomainNet. We present FAA ($\uparrow$) across different task scales (5, 10, and 15 tasks).} 
    \renewcommand\arraystretch{1.2}
    \setlength\tabcolsep{4pt}
    \small{
    \resizebox{0.38\textwidth}{!}{ 
    \begin{tabular}{l|ccc}
     \hline
      \multirow{2}{*}{Method} & \multicolumn{3}{c}{DomainNet} \\
        & 5-task & 10-task & 15-task \\
        \hline
       L2P  & 69.58 & 66.82 & 64.15 \\ 
       DualPrompt  & 70.73 & 68.11 & 65.48 \\ 
       S-Prompt++  & 72.35 & 69.54 & 66.88 \\ 
       CODA-Prompt  & 73.24 & 70.62 & 68.03 \\ 
       HiDe-Prompt & \textbf{78.92} & 75.68 & 72.85 \\ 
       Ours & 78.65  & \textbf{76.54} & \textbf{73.91} \\
       \hline
    \end{tabular}
    } 
}
    \label{table:domainnet_scaling}
\end{table}

Table~\ref{table:domainnet_scaling} presents a similar trend on the DomainNet benchmark. In this setting, HiDe-Prompt obtains the best performance in the 5-task scenario where the pressure for knowledge reuse is lower. However, as the task length increases, its performance degrades more significantly compared to our approach. 

Overall, these results across both benchmarks support the conclusion that our routing mechanism facilitates more efficient training. By accurately distinguishing between novel information and recurring concepts, the model effectively balances parameter isolation with knowledge reuse, leading to superior scalability over extended horizons.

\noindent\textbf{Robustness on Mixed Data Streams.}
To further evaluate the flexibility of our routing mechanism, we utilize the mixed-dataset protocol \cite{ke2021adapting}, denoted as M-EMNIST and M-CifarCeleb. In this setting, the task sequence is heterogeneous: some tasks introduce entirely new classes, while others present samples from the same overall distribution as previously encountered tasks, such as handwritten samples from a previously seen author or photos of a known individual. 

\begin{table}[h]
\small{
\begin{tabular}{l|cccccc}
\hline
\multicolumn{1}{c}{\multirow{2}{*}{Method}} & \multicolumn{3}{c}{M-EMNIST} & \multicolumn{3}{c}{M-CifarCeleb} \\ \cline{2-7} 
\multicolumn{1}{c}{} & \multicolumn{1}{c}{FAA} & \multicolumn{1}{c}{CA} & \multicolumn{1}{c}{FM} & \multicolumn{1}{c}{FAA} & \multicolumn{1}{c}{CA} & \multicolumn{1}{c}{FM } \\ \hline
L2P          & 93.12 & 94.50 & 2.11 & 72.44 & 76.88 & 6.40 \\
DualPrompt   & 95.88 & 96.92 & 1.85 & 78.10 & 81.55 & 5.11 \\
CODA-Prompt  & 95.50 & 97.10 & 1.90 & 77.65 & 82.40 & 5.45 \\
S-Prompt++     & 96.12 & 97.25 & 1.40 & 76.90 & 80.22 & 4.92 \\
HiDe-Prompt  & 98.45 & 98.90 & 0.75 & 86.55 & 88.92 & 2.85 \\ 
Ours  & \textbf{99.31} & \textbf{99.55} & \textbf{0.48} & \textbf{90.11} & \textbf{92.70} & \textbf{2.38} \\ \hline
\end{tabular}
}
\caption{Evaluation on task sequences with a mixture of similar and dissimilar tasks. }
\label{table:mixed_datasets}
\end{table}

As shown in Table~\ref{table:mixed_datasets}, our method achieves state-of-the-art performance in these mixed environments. The performance gap suggests that our routing mechanism is better equipped to distinguish between novelty and recurrence within the stream. Specifically, while baseline methods are typically optimized for disjoint boundaries and may treat every task as a new learning problem, our approach appears to correctly route samples from recurring distributions back to established prompts for refinement. This behavior likely prevents the parameter redundancy observed in boundary-based methods, which tend to allocate unnecessary capacity when a known concept reappears under a different task label. These results indicate that by decoupling prompt selection from the chronological task index, the model can more effectively balance the reuse of existing knowledge with the isolation of new information.

\noindent\textbf{Verification on Standard Benchmarks.} Finally, we evaluate our method on the standard Class-Incremental Learning (CIL) benchmarks, Split CIFAR-100 and Split ImageNet-R. In these protocols, the task sequence is strictly disjoint and sequential, with ample training data for each task.

\begin{table*}[t]
    \centering
    \caption{Overall performance of continual learning. We present the final average accuracy (FAA), cumulative average accuracy (CAA) and final forgetting measure (FFM) with $\pm$ standard deviation, over three runs of different random seeds and task splits.} 
    \small{
    \resizebox{0.9\textwidth}{!}{ 
    \begin{tabular}{l|ccc|ccc}
     \hline
        \multirow{2}{*}{Method} & \multicolumn{3}{c|}{Split CIFAR-100} & \multicolumn{3}{c}{Split ImageNet-R} \\
        & FAA ($\uparrow$) & CAA ($\uparrow$) & FFM ($\downarrow$) & FAA ($\uparrow$) & CAA ($\uparrow$) & FFM ($\downarrow$)\\
        \hline
       L2P  &83.06 \small{$\pm 0.17$}&88.25 \small{$\pm 0.01$} &6.58 \small{$\pm 0.40$} &63.65 \small{$\pm 0.12$} &67.25 \small{$\pm 0.02$} &7.51 \small{$\pm 0.17$} \\ 
       DualPrompt &86.60 \small{$\pm 0.19$} &90.64 \small{$\pm 0.01$} &4.45 \small{$\pm 0.16$} &68.79 \small{$\pm 0.31$} &71.96 \small{$\pm 0.04$} &4.49 \small{$\pm 0.14$} \\ 
       S-Prompt++  &88.81 \small{$\pm 0.18$} &92.25 \small{$\pm 0.03$} &3.87 \small{$\pm 0.05$} & 69.68 \small{$\pm 0.12$}&72.50 \small{$\pm 0.04$} &3.29 \small{$\pm 0.05$} \\
       CODA-Prompt &86.94 \small{$\pm 0.63$} &91.57 \small{$\pm 0.75$} &4.04 \small{$\pm 0.18$} & 70.03 \small{$\pm 0.47$}& 74.26 \small{$\pm 0.24$} &5.17 \small{$\pm 0.22$}\\
       HiDe-Prompt  &\textbf{92.61} \small{$\pm 0.28$} & \textbf{94.03} \small{$\pm 0.01$} & 3.16 \small{$\pm 0.10$} & 75.06 \small{$\pm 0.12$} & 76.60 \small{$\pm $0.01} & 2.17 \small{$\pm 0.19$} \\ 
       Ours & 92.53 \small{$\pm 0.22$} & 93.80 \small{$\pm 0.03$} & \textbf{3.18} \small{$\pm 0.12$} & \textbf{75.10} \small{$\pm 0.15$} & \textbf{78.19} \small{$\pm 0.02$} & \textbf{2.15} \small{$\pm 0.17$} \\
       \hline
    \end{tabular}
    } }
    \label{table:cifar_imagenet}
\end{table*}

As presented in Table~\ref{table:cifar_imagenet}, our method achieves performance that is competitive with the current state-of-the-art HiDe-prompt. While this benchmark does not challenge the sample efficiency or mixed-sequence resilience as our previous studies, it serves as a verification for the stability of our method. 

\subsection{Ablation Experiments}
\begin{figure}[h]
    \centering
\begin{subfigure}[b]{\linewidth}
  \includegraphics[width=1\linewidth, trim={0 2.7cm 0 0}]{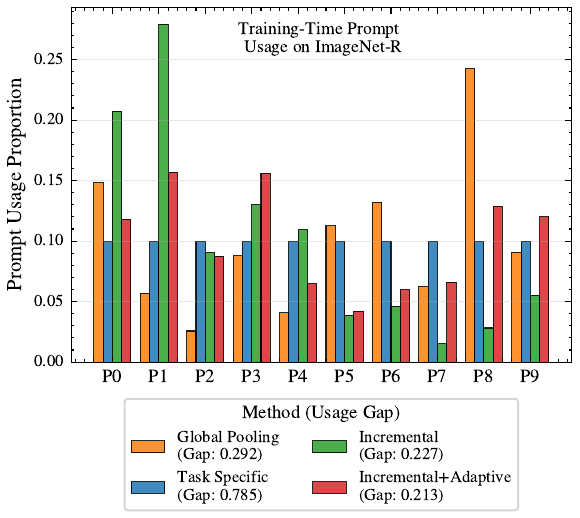}
  \label{fig:Ng1} 
\end{subfigure}

\medskip 
\begin{subfigure}[b]{\linewidth}
  \includegraphics[width=1\linewidth, trim={0 0.8cm 0 0}]{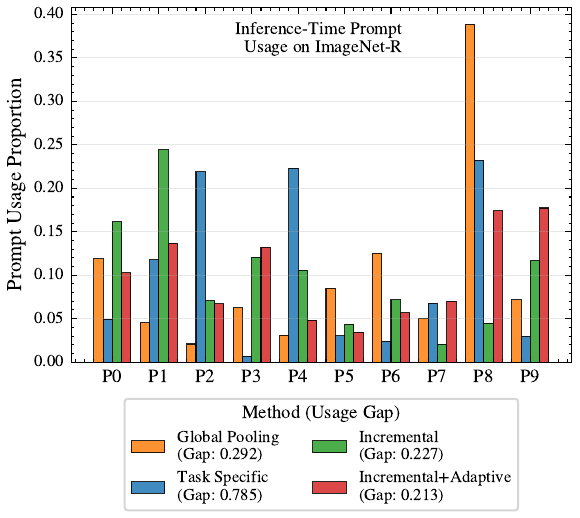}
  \label{fig:Ng2}
\end{subfigure}

\caption{Comparison of prompt allocation during training (top) and inference time (bottom) on ImageNet-R (10-Tasks). }
    \label{fig:prompt_usage}
\end{figure}

To investigate the underlying mechanism driving the performance gains observed in the preceding sections, we visualize the prompt selection statistics during both training and inference in Figure~\ref{fig:prompt_usage}.

During training, task-specific methods exhibit a uniform distribution of prompt updates, as routing is determined rigidly by task ID. In contrast, global pooling baselines show sharper concentration on certain prompts, while our adaptive routing demonstrates a more uniform distribution with selective peaks at earlier prompts, indicating effective knowledge reuse.

At inference time, the global pooling and our proposed routing methods largely maintain their training distributions. However, task-specific allocation reveals a critical structural inefficiency: the selection frequency becomes significantly sparse, with many prompts that were uniformly updated during training now rarely retrieved for prediction. This misalignment between training and inference suggests that the value of the training samples was not fully utilized, explaining the lower parameter efficiency observed in low-data regimes.

We formalize this phenomenon by defining the \textit{Usage Gap} ($\Delta_{\text{usage}}$) as the $L_1$ distance between the proportion of samples assigned to each prompt during training ($p_{\text{train}}$) and inference ($p_{\text{val}}$):
\begin{equation}
\Delta_{\text{usage}} = \sum_{i=1}^{|\calP|} |p_{\text{train}, i} - p_{\text{val}, i}|,
\end{equation}

where $p \in [0,1]^{|\mathcal{P}|}$ represents the vector of assignment proportions across the entire prompt set $\mathcal{P}$. Task-specific methods exhibit a high $\Delta_{\text{usage}}$, while our adaptive routing maintains the closest alignment.

While global pooling achieves lower $\Delta_{\text{usage}}$ than task-specific allocation, it over-utilizes a small subset of prompts to represent data from multiple disparate tasks. This capacity collapse leads to representational interference and performance decay over long horizons, particularly when task heterogeneity is high. Our adaptive routing avoids both extremes: it maintains alignment between training and inference while distributing capacity to prevent interference, ensuring that the model’s learned representations remain both accessible and task-appropriate. This structural efficiency drives the improvements in scalability and robustness demonstrated across the previous experiments.

\section{Conclusion}
In this work, we addressed the challenges of Continual Learning under per-task data scarcity and unstructured task overlap. We investigated the inherent trade-offs between global pooling and task-specific training, specifically the delicate balance between facilitating knowledge sharing and avoiding inter-task interference. To navigate this, we developed an adaptive mixture-of-experts framework based on incremental global pooling with instance-wise prompt masking.

Our empirical evaluation demonstrates that this approach yields consistent improvements over state-of-the-art baselines, particularly in low-data regimes,  extended task streams, and scenarios involving a complex mixture of similar and dissimilar tasks. These results indicate the efficacy of knowledge sharing with out-of-distribution awareness for efficient continual learning.

Future work could investigate alternative OOD detection mechanisms tailored for continual streams or explore tighter integration of distribution-aware metrics directly into the query-key matching process to further refine parameter efficiency.

\bibliographystyle{ACM-Reference-Format}
\bibliography{ref}

\end{document}